\documentclass[letterpaper, 10 pt, journal, twoside]{ieeetran}

\newcommand{\abs}[1]{|#1|}

\usepackage{amsmath,amsfonts}
\usepackage{algorithmic}
\usepackage{algorithm}
\usepackage{array}
\usepackage[caption=false,font=normalsize,labelfont=sf,textfont=sf]{subfig}
\usepackage{textcomp}
\usepackage{stfloats}
\usepackage{url}
\usepackage{colortbl}
\usepackage{verbatim}
\usepackage{graphicx}
\usepackage{mathrsfs}
\usepackage{array}
\usepackage{color}
\usepackage{cite}
\usepackage{tikz}
\usepackage{multirow}
\usepackage{pifont}
\newcommand{\xmark}{\ding{55}}%
\title{Meta-RangeSeg: LiDAR Sequence Semantic Segmentation Using Multiple Feature Aggregation}

\author{
Song Wang$^{1}$, Jianke~Zhu$^{1}$,~\IEEEmembership{Senior~Member,~IEEE}, and Ruixiang Zhang$^{2}$

\thanks{$^{1}$Song~Wang and Jianke~Zhu are with the College of Computer Science, Zhejiang University, Hangzhou 310027, China, and also with the Alibaba-Zhejiang University
Joint Research Institute of Frontier Technologies, Hangzhou, China. Jianke~Zhu is the corresponding author. (email: songw@zju.edu.cn; jkzhu@zju.edu.cn). }
\thanks{$^{2}$Ruixiang~Zhang is with Hikvision Research Institute, Hangzhou 310051, China. (email: zrxisgood@gmail.com).}

}

\begin{document}
\maketitle

\markboth{IEEE ROBOTICS AND AUTOMATION LETTERS. PREPRINT VERSION.}
{Wang \MakeLowercase{\textit{et al.}}: Meta-RangeSeg: LiDAR Sequence Semantic Segmentation}

\begin{abstract}
LiDAR sensor is essential to the perception system in autonomous vehicles and intelligent robots. To fulfill the real-time requirements in real-world applications, it is necessary to efficiently segment the LiDAR scans. Most of previous approaches directly project 3D point cloud onto the 2D spherical range image so that they can make use of the efficient 2D convolutional operations for image segmentation. Although having achieved the encouraging results, the neighborhood information is not well-preserved in the spherical projection. Moreover, the temporal information is not taken into consideration in the single scan segmentation task. To tackle these problems, we propose a novel approach to semantic segmentation for LiDAR sequences named Meta-RangeSeg, where a new range residual image representation is introduced to capture the spatial-temporal information. Specifically, Meta-Kernel is employed to extract the meta features, which reduces the inconsistency between the 2D range image coordinates input and 3D Cartesian coordinates output. An efficient U-Net backbone is used to obtain the multi-scale features. Furthermore, Feature Aggregation Module (FAM) strengthens the role of range channel and aggregates features at different levels. We have conducted extensive experiments for performance evaluation on SemanticKITTI and SemanticPOSS. The promising results show that our proposed Meta-RangeSeg method is more efficient and effective than the existing approaches. Our full implementation is publicly available at \url{https://github.com/songw-zju/Meta-RangeSeg}.
\end{abstract}


\section{Introduction}
\label{sec:intro}

\IEEEPARstart{L}{iDAR} can accurately measure the range by taking advantage of its active sensor, which plays an increasingly important role in the perception system of modern autonomous vehicles and robotics. Due to the characteristics of disorder and irregularity in point cloud, it is challenging to perform scene understanding on LiDAR sequences.

LiDAR semantic segmentation aims to estimate the labels for each point, which is the key to understand the surroundings for the perception system. During past decade, extensive research efforts have been devoted to this task. Point-based methods~\cite{qi2017pointnet, qi2017pointnet++, hu2020randla} directly extract features from the raw output of LiDAR sensor. However, point convolution is usually computational intensive. To address this issue, projection-based methods~\cite{wu2019squeezesegv2,milioto2019rangenet++,cortinhal2020salsanext} and voxel-based approaches~\cite{graham20183d, yan2021sparse, zhu2021cylindrical} convert the irregular raw point cloud data into regular grid representations so that the conventional convolutional layer for image can be employed. Nevertheless, they fail to preserve the original neighborhood relationship. In practice, the hybrid methods~\cite{zhang2020deep, tang2020searching, xu2021rpvnet} fuse two or more of the above feature representations, which can obtain better results. Unfortunately, this incurs the extra computational load. 

\begin{figure}
\centering
\setlength{\abovecaptionskip}{-0.2cm}
\includegraphics[width=0.9\linewidth,height=85mm]{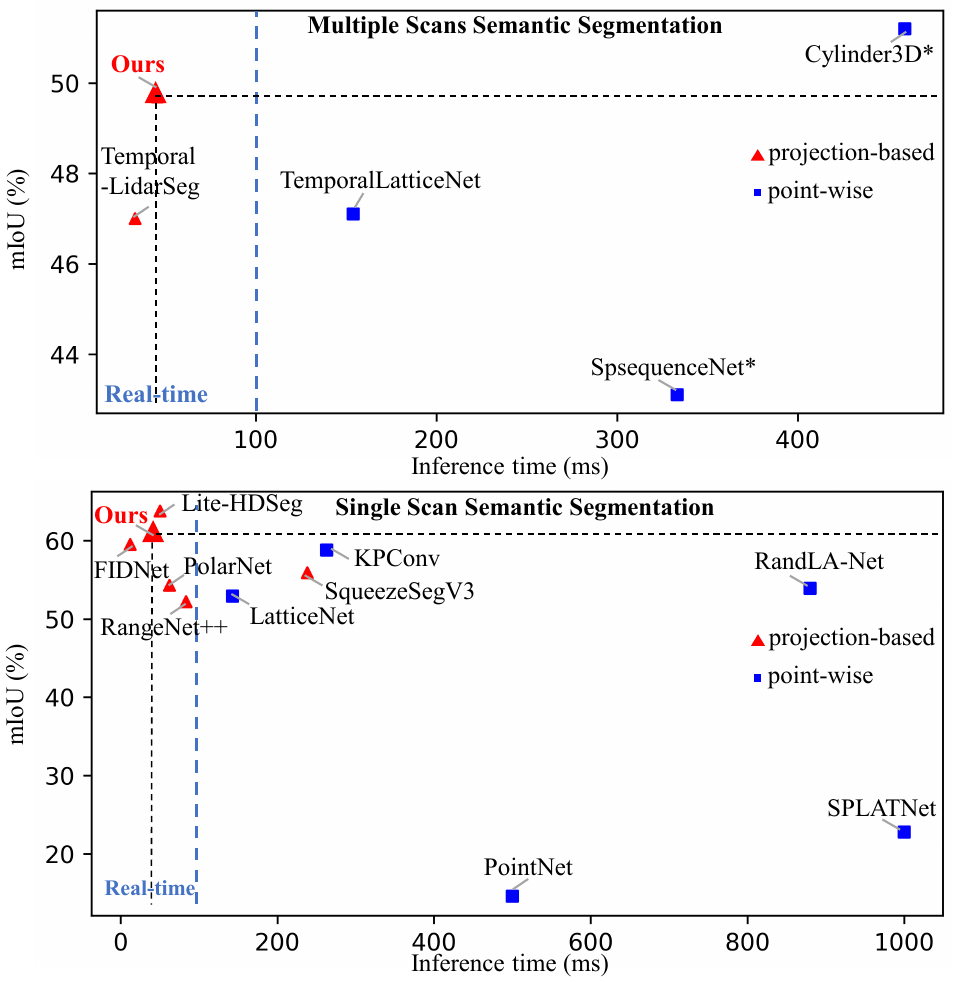}
    \caption{Accuracy vs. inference time. $*$ denotes the results reproduced from the original implementation. Our presented Meta-RangeSeg method obtains the promising results on both multiple scans and single scan semantic segmentation in SemanticKITTI benchmark~\cite{behley2019semantickitti} and runs at real-time.}
    \label{fig:first}
    \vspace{-6mm}
\end{figure}

Generally, scene analysis for autonomous driving is conducted within a sequence of LiDAR scans. Most of previous approaches only take into account of single frame, where the important temporal information is usually ignored. Moreover, some methods~\cite{sun2021pointmoseg, shi2020spsequencenet, zhu2021cylindrical} aim to deal with multiple scans simultaneously. This may lead to information redundancy and slow inference speed, as shown in Fig.~\ref{fig:first}.

To tackle the above challenges, we propose a novel approach to semantic segmentation on LiDAR sequences named Meta-RangeSeg. To this end, a new range residual image representation is introduced to capture the spatial-temporal information.
In contrast to the direct fusion methods, our proposed range residual image efficiently represents multi-frame point cloud information, which can improve the accuracy and the speed of training and inference under the limited computing resources. Since the range residual image obtained from spherical projection may not effectively capture the local geometric structures, we take advantage of the Meta-Kernel operator~\cite{fan2021rangedet} to extract the meta features by dynamically learning the weights from the relative Cartesian coordinates and range values. 
Thus, it reduces the inconsistency between the 2D range image coordinates input and Cartesian coordinates output. Moreover, an efficient U-Net backbone \cite{ronneberger2015u} is used to obtain the multi-scale features. Feature Aggregation Module (FAM) aggregates the meta features and multi-scale features with range guided information. We have conducted extensive experiments for performance evaluation on SemanticKITTI~\cite{behley2019semantickitti} and SemanticPOSS~\cite{pan2020semanticposs} datasets. The promising results show that our proposed method is more efficient and effective than the existing approaches. 

In summary, the main contributions of this paper are: 1) a novel framework for semantic segmentation on LiDAR sequences by taking advantage of range residual image, which is able to capture the spatial-temporal information efficiently; 2) an effective Meta-Kernel based feature extraction method for LiDAR semantic segmentation; 3) a Feature Aggregation Module (FAM) to aggregate features at various scales and levels for range-based object segmentation; 4) experiments on SemanticKITTI and SemanticPOSS benchmark show that our proposed approach is promising.

\section{Related Work}\label{sec:related}
With the prevalence of autonomous driving, a surge of research efforts have been spent on semantic scene understanding~\cite{behley2019semantickitti,pan2020semanticposs,hu2022sensaturban}. In this work, we focus on the task of semantic segmentation using LiDAR scans~\cite{behley2019semantickitti,pan2020semanticposs}. Generally, most of existing studies on LiDAR semantic segmentation can be categorized into four groups according to the different feature representations, including point, projection-based image, voxel and hybird.

Point-based methods directly extract features from the raw point cloud data, which are able to preserve the 3D spatial structure information. Due to irregularity of point cloud data, it is challenging to design the efficient neural network layer for it. Qi \textit{et al.}~\cite{qi2017pointnet} extract the deep features on point cloud by the shared Multi-Layer Perceptrons (MLP) for classification and segmentation. The subsequent series of works~\cite{qi2017pointnet++,thomas2019kpconv} try to address the limitation in extracting local features, which obtain the encouraging results on the indoor semantic segmentation. The main showstopper for these approaches is their high computational cost and memory consumption, which hinders them from the large-scale outdoor driving scenarios. One remedy is to reduce their time complexity and information loss by randomly sampling and local feature aggregation~\cite{hu2020randla}. Despite of its efficiency on the large scenes, there is noticeable performance drop due to sub-sampling.

Voxel representation is able to make use of the 3D convolution neural network that can effectively solve the irregularity problem of point cloud. The regular 3D dense convolution for semantic segmentation requires the huge memory and heavy computational power for the fine resolution, which limit their capability of processing the large scale outdoor LiDAR scans. To this end, the sparse convolution~\cite{graham20183d, yan2021sparse} is employed to reduce the computational cost. Zhu \textit{et al.}~\cite{zhu2021cylindrical} propose a cylindrical voxel division method with asymmetric convolution based on LiDAR point cloud distribution.

By projecting 3D point cloud onto 2D space, range image is a promising representation, which can take advantage of a large amount of advanced layers for image feature extraction with fast training and inference. To account for the mechanism of LiDAR scanning, most of existing LiDAR semantic segmentation approaches~\cite{milioto2019rangenet++,cortinhal2020salsanext,xu2020squeezesegv3, alonso20203d} make use of spherical projection to obtain range images. Besides range view (RV), Zhang \textit{et al.}~\cite{zhang2020polarnet} and Wen \textit{et al.}~\cite{wen2022hybrid} employ a bird's-eye view (BEV) for semantic segmentation. Some studies~\cite{liong2020amvnet,gerdzhev2021tornado} combine these two projection methods in order to achieve more accurate segmentation results. However, this will lead to the extra memory consumption and computational overhead. Moreover, directly fusing two different projections ignores the underlying geometric structure of LiDAR scan.

The hybrid approaches intend to fuse the different feature representations for better LiDAR semantic segmentation. Zhang \textit{et al.}~\cite{zhang2020deep} propose a point-voxel feature aggregation module that aggregates features among neighborhood voxels and produces point-wise prediction. Thereby, it is able to avoid the time-consuming neighborhood search while achieving the encouraging results on outdoor LiDAR data. Tang \textit{et al.}~\cite{tang2020searching} present an efficient point-voxel fusion pipeline. Voxels provide the coarse-grained local features, and points preserve the fine-grained geometric features through a simple MLP. Xu \textit{et al.}~\cite{xu2021rpvnet} fuse three different feature representations, including point, range image and voxel, which achieve the promising fusion results by interacting features at various stages. 
Besides, Zhuang \textit{et al.}~\cite{zhuang2021perception} try to fuse the multiple modalities like image and point cloud.

Most of existing approaches perform the LiDAR semantic segmentation on single scan, where the temporal information is usually neglected. There are only few methods focusing on the multiple scans task. Shi \textit{et al.}~\cite{shi2020spsequencenet} employ a voxel-based 3D sparse convolutional network to fuse local information from the previous and current frames through local interpolation, which only make use of the two consecutive scans. Duerr \textit{et al.} ~\cite{duerr2020lidar} propose a novel recurrent segmentation framework using range images, which recursively aggregate the features of previous scans in order to exploit the short term temporal dependencies. 
Schütt \textit{et al.} ~\cite{schutt2022abstract} extend the original LatticeNet~\cite{rosu2019latticenet} with a novel abstract flow module for temporal semantic segmentation.
In~\cite{zhu2021cylindrical}, accumulating point clouds in 3D space is adopted for multiple scans segmentation, whose memory consumption and computational time increase linearly with the total number of scans per input model. In this paper, we introduce an efficient range residual image representation, where the effective features can be extracted by Meta-Kernel and U-Net backbone.

\section{Meta-RangeSeg \\ for LiDAR Semantic Segmentation}
\begin{figure*}
\centering
\setlength{\abovecaptionskip}{-0.2cm}
\includegraphics[width=0.9\linewidth,height=72mm]{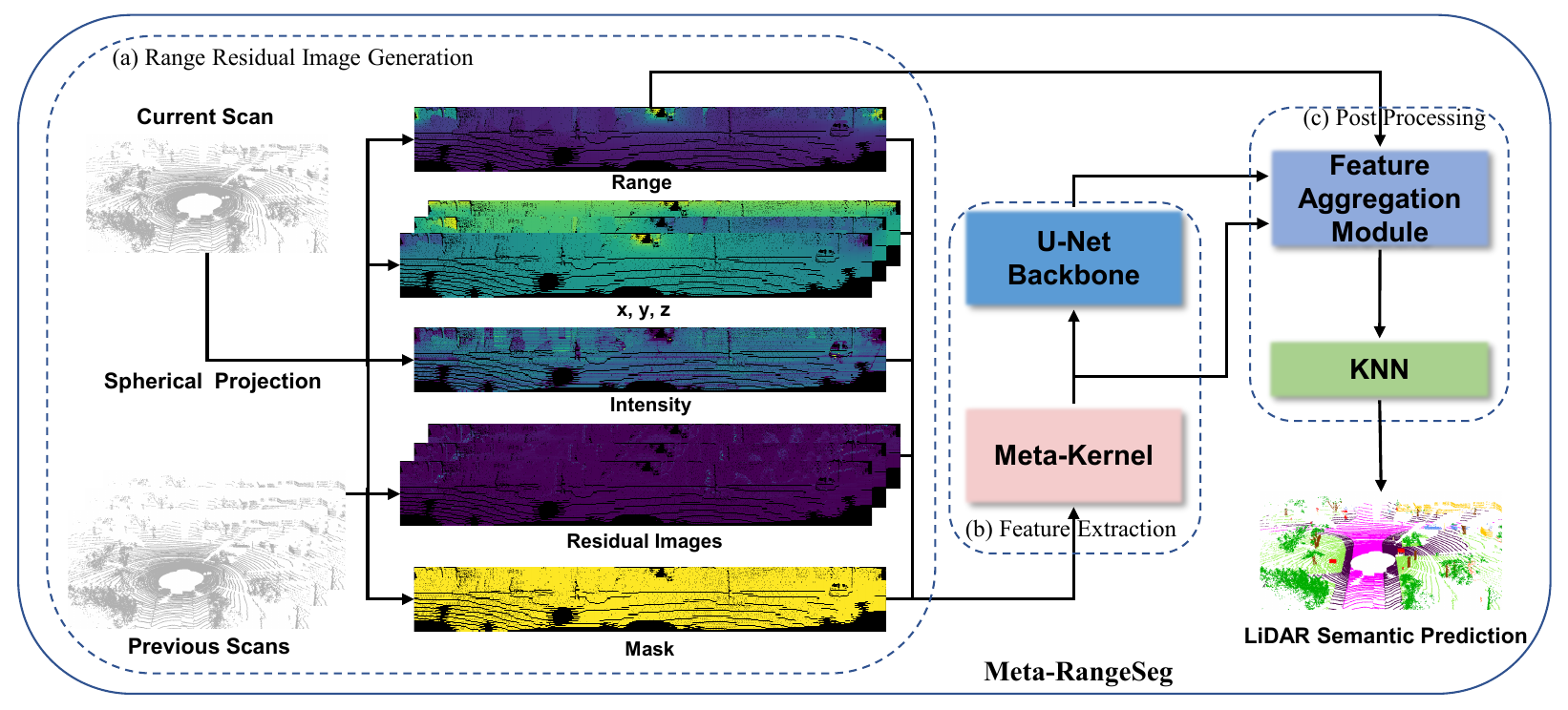}
    \caption{Our proposed Meta-RangeSeg framework. (a) Calculate range residual image with nine channels capturing spatial and motion information of raw data. (b) Extract meta features by Meta-Kernel and obtain multi-scale features via U-Net backbone.
    (c) Aggregate features and get semantic labels in 3D space.}
    \label{fig:second}
    \vspace{-4mm}
\end{figure*}

\label{sec:main}

In this section, we present an efficient neural network Meta-RangeSeg for LiDAR semantic segmentation on multiple scans.

\subsection{Overview}
In this paper, we aim to predict the semantic labels from the consecutive LiDAR sequences. Unlike the conventional approaches transforming the sequential point cloud into global coordinates~\cite{shi2020spsequencenet, zhu2021cylindrical} in 3D space, we suggest a novel approach named Meta-RangeSeg to efficiently process multiple scans in range view for the subsequent feature extraction.

As shown in Fig.~\ref{fig:second}, our proposed network takes advantage of the range residual image with nine channels built from the current scan and previous ones. Then, the meta features are extracted by a Meta-Kernel block, and the multi-scale features are obtained via a U-Net network. Finally, we get the semantic labels for raw data by post-processing the aggregated features. In the following, we will give the detailed description of range residual image and our network architecture.

\subsection{Range Residual Image}
The traditional range image is a multi-channel pseudo image obtained by spherical projection of the LiDAR point cloud. Each channel represents range ($r$), $x$, $y$, $z$ and remission ($e$) sequentially. Range image representation has the advantage of using the effective 2D operations for fast training and inference. To this end, we map the scattered LiDAR points into their corresponding 2D spherical coordinates through a mapping function $\mathbb{R}^3 \rightarrow \mathbb{R}^2$ as below: 
\begin{align}
	\left( \begin{array}{c} u \vspace{0.0em} \\ v \end{array}\right) & = \left(\begin{array}{cc} \frac{1}{2}\left[1-\arctan(y, x) \, \pi^{-1}\right]~\,~W   \vspace{0.5em} \\
			\left[1 - \left(\arcsin(z\, r^{-1}) +f_{up}\right) {f}^{-1}\right] \, H\end{array} \right), \label{eq:projection}
\end{align}
where $(H, W)$ are the height and width of the range image, and $r~{=}\sqrt{x^{2}+y^{2}+z^{2}}$ is the range value of the point in 3D space. $(u,v)$ are the image coordinates under range view. ${f}~{=}~{f}_{{up}}~{+}~{f}_{{down}}$ is the vertical field-of-view of the LiDAR. 

Motivated by the residual image feature for segmenting moving objects in video analysis~\cite{wang2018temporal,chen2021moving}, we introduce it into the task of semantic segmentation on LiDAR sequences to capture the temporal information. Based on the range images from current scan and previous frames using spherical projection, the input of our proposed neural network is made of range images and their residuals, as illustrated in Fig.~\ref{fig:second}. Specifically, it is a range residual image with the size of $9 \times H \times W$, where each pixel $(u,v)$ contains a vector $(r, x, y, z, e, d_{1}, d_{2}, d_{3}, m)$.  The mask $m$ indicates whether the pixel position is a projected point or not. $d_{k} (k = 1, 2, 3..)$ is from the residual image  that calculates the range differences between the previous $k^\text{th}$ scan and the current one.  We use the last three LiDAR scans in our implementation.

To effectively fuse different scans, residual image is calculated in the following operations. We firstly transform the point clouds of previous frames into the coordinate of current frame. Then the transformed point clouds are projected into the range view with Eq.~\eqref{eq:projection}. Finally, the residual value $d_{k}$ for each pixel is obtained by calculating the absolute differences between the ranges of current scan and the transformed one with normalization as
\begin{align}
	d_{k} =  \frac{\abs{r - r_{k}}}{r},
\end{align}
where $r$ is the range value in current scan, and $r_{k}$ is the corresponding one from the transformed point cloud projected at the same image coordinate.

\subsection{Feature Extraction}
To facilitate the effective LiDAR semantic segmentation, we design a feature extraction module that consists of a Meta-Kernel block and a U-Net backbone, as shown in Fig.~\ref{fig:second}.

In our empirical study, the conventional convolution operations do not perform well on the range residual image. To address this issue, we take advantage of the Meta-Kernel block to extract the meta features by dynamically learning the weights from the relative Cartesian coordinates and range values. As in~\cite{fan2021rangedet}, the Meta-Kernel is designed to effectively locate objects in LiDAR scans by exploiting the geometric information from the Cartesian coordinates. In this paper, we employ it to capture the spatial and temporal information for semantic segmentation.

\begin{figure}
 \centering
 \setlength{\abovecaptionskip}{-0.2cm}
\includegraphics[width=0.8\linewidth,height=70mm]{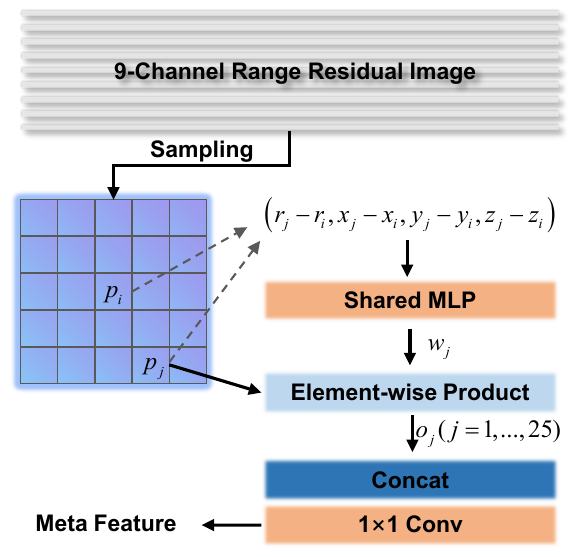}
    \caption{Illustration of Meta-Kernel. We sample the range residual image using a $5 \times 5$ sliding window, and learn weights dynamically from the relative Cartesian coordinates and range value to extract meta features.  }
    \label{fig:third}
    \vspace{-6mm}
\end{figure}

In order to achieve a larger receptive field on the range residual image for Meta-Kernel, we enlarge the size of sliding window into $5 \times 5$. Therefore, we can get the relative Cartesian coordinates of 25 neighbors $\bold{p}_{j}(x_j, y_j, z_j)$ for the center $\bold{p}_{i}(x_i, y_i, z_i)$ : $\left(x_{j}-x_{i}, y_{j}-y_{i}, z_{j}-z_{i}\right)$. More importantly, the range difference is added to enhance the perception of spatial information. Furthermore, a shared MLP is employed to generate 25 weight vectors from the input $\left(r_j-r_i, x_{j}-x_{i}, y_{j}-y_{i}, z_{j}-z_{i}\right)$. We multiply the learned weight vectors $\bold{w}_j$ element-wisely with the channel information $(r_j, x_j, y_j, z_j, e_j, d_{1,j}, d_{2,j}, d_{3,j}, m_j)$ at its corresponding position within the sliding window. Finally, a $1 \times 1$ convolution is used to obtain the meta features, which aggregates the information from different channels and different sampling locations. From the above all, the whole process of Meta-Kernel block is summarized into Fig.~\ref{fig:third}.

We obtain the multi-scale features through the U-Net backbone that is an encoder-decoder architecture commonly used in semantic segmentation~\cite{ronneberger2015u,cortinhal2020salsanext}. Specifically, we firstly employ four down-sampling layers to extract the features of different scales from meta features, and then restore the original resolution through four up-sampling layers. Moreover, skip connection is adopted to assist in reconstructing high-resolution semantic information.

\subsection{Post Processing}
\begin{figure}
 \centering
 \setlength{\abovecaptionskip}{-0.2cm}
\includegraphics[width=0.8 \linewidth,height=90mm]{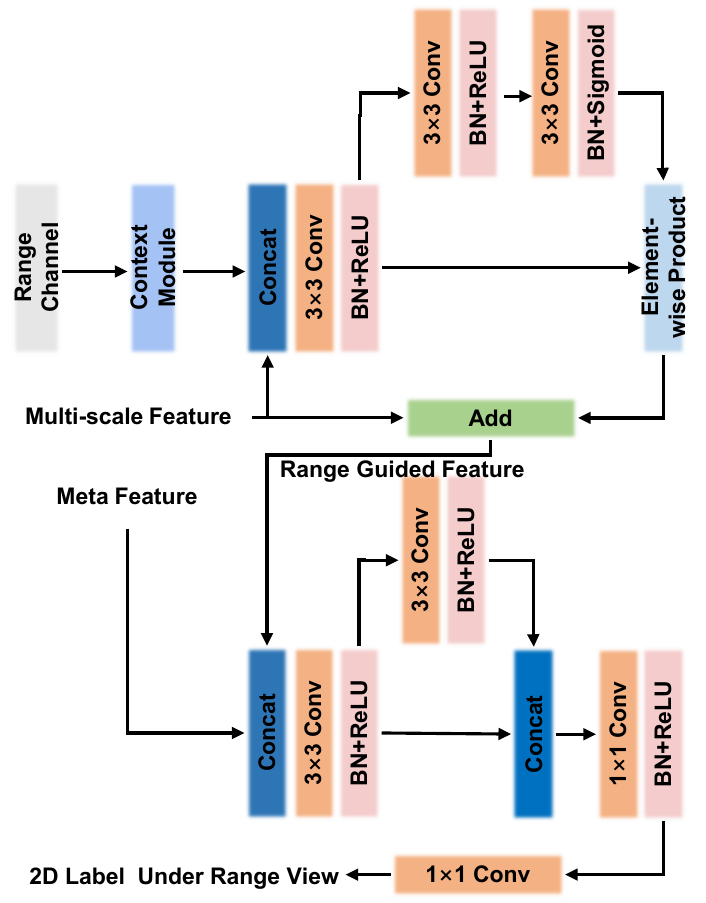}
    \caption{Illustration of Feature Aggregation Module (FAM). We fistly extract the features of the range channel separately, and then fuse the multi-scale and the range features to obtain the range-guided features. Finally, the meta and range-guided features are aggregated via skip connection, and the 2D label under range view is obtained through $1 \times 1$ convolution layer.}
    \label{fig:fourth}
    \vspace{-6mm}
\end{figure}

Once the meta and multi-scale features are obtained, we perform feature aggregation to predict the labels from range perspective. To this end, the Feature Aggregation Module is designed to make use of range information for object segmentation in the different ranges by aggregating features at various scales and levels, as shown in Fig.~\ref{fig:fourth}.

As the $x$, $y$ and $z$ information has been encoded in the horizontal and vertical coordinates under range view, the range channel is the most important one in range residual image. Thus, they are extracted separately.
It is further fed into the context module~\cite{cortinhal2020salsanext} to capture the global features with more detailed range context. Then, we fuse the range features with multi-scale features through an attention fusion layer. To this extent, the range-guided features are able to aggregate the range context information and semantic features. Furthermore, the meta features are reused through skip connections. We concatenate the meta features with the range-guided features. Afterwards, the final features are fused through the convolution and residual connection. Thus, we can obtain the 2D semantic labels under the range view via a $1 \times 1$ convolution layer.

To estimate the labels in 3D space from 2D predictions in range view, we employ $k$-Nearest Neighborhood in the post-processing stage. During the spherical projection, there may be multiple points projected into the same grid. We sort the points according to their ranges, and the characteristics of the closer points within the range shall prevail. When recovering 3D information from 2D range residual image, we need to supplement the features of those missing points. In $k$-NN , the label of each point is jointly determined by its $k$ closest points. Instead of using Euclidean distance, range is employed as the similarity measure so that we can efficiently process data using the sliding windows in 2D space. As described in~\cite{milioto2019rangenet++}, the $k$-nearest points within a window can represent the distribution in 3D space very well. In our implementation, we set $k = 5$ with a $ 7 \times 7$ sliding window.

\subsection{Loss Function}
To facilitate the effective semantic segmentation, we train the proposed neural network by the loss function $\mathcal{L}$ with three different terms as follows:
\begin{align}
    {\mathcal{L}}=w_1{\mathcal{L}}_{wce}+w_2{\mathcal{L}}_{ls}+ w_3{\mathcal{L}}_{bd},
\end{align}
where ${\mathcal{L}}_{wce}$ is the weighted cross-entropy loss, and ${\mathcal{L}}_{ls}$ is the Lov\'asz loss. ${\mathcal{L}}_{bd}$ is the Boundary loss. $w_1$, $w_2$ and $w_3$ are the weights with respect to each term. In our implementation, we set $w_1 = 1$, $w_2 = 1.5$ and $w_3 = 1$, empirically.

To account for the multi-class segmentation problem, the weighted cross-entropy loss ${\mathcal{L}}_{wce}$~\cite{zhang2018generalized} is employed to maximize the prediction accuracy for point labels, which is able to balance the distributions among different classes. It weights the cross-entropy loss of every class with the corresponding frequency $f_i$ as 
\begin{equation}
\mathcal{L}_{wce}(y, \hat{y})=-\sum_{i} \frac{1}{\sqrt{f_{i}}}  p\left(y_{i}\right) \log \left(p\left(\hat{y}_{i}\right)\right),
\end{equation}
where $y_i$ represents the ground truth, and $\hat{y}_i$ is prediction.

The Lov\'asz loss ${\mathcal{L}}_{ls}$~\cite{berman2018lovasz} is used to maximize the intersection-over-union (IoU) score that is commonly used to in performance evaluation on semantic segmentation. Since IoU is discrete and indifferentiable, it needs to be optimized using a derivable surrogate function. We define a vector of pixel errors $\bold{m}(c)$ of each pixel $i$ on class $c$ with its predicted probability $f_{i}(c) \in[0,1]$ and ground truth label $y_{i}(c) \in\{-1,1\}$:
\begin{equation}
m_{i}(c)= \begin{cases}1-f_{i}(c) & \text { if } c=y_{i}(c),  \\ f_{i}(c) & \text { otherwise }\end{cases}
\end{equation}
We use the Lov\'asz extension \cite{berman2018lovasz} for the vector of errors $\bold{m}(c)$ to construct the loss $\overline{\Delta_{J_{c}}}$ surrogate to $\Delta_{J_{c}}$. Then, ${\mathcal{L}}_{ls}$ can be formulated as below:
\begin{equation}
\mathcal{L}_{l s}=\frac{1}{|C|} \sum_{c \in C} \overline{\Delta_{J_{c}}}(\bold{m}(c)),
\end{equation}
where $|C|$ denotes the total number of classes. 

As suggested in~\cite{bokhovkin2019boundary}, the pixel-level loss function like cross-entropy may not effectively handle the complex boundaries between different classes in remote sensing images that have a wide range and low contrast. To emphasize the boundaries between different objects, we adopt the boundary loss function ${\mathcal{L}}_{bd}$~\cite{bokhovkin2019boundary} for LiDAR semantic segmentation. Given the extracted boundary image $y^{b}$ for ground truth $y$ and $\hat{y}^{b}$ for the predicted result $\hat{y}$ in the range view, ${\mathcal{L}}_{bd}$ is defined as below:
\begin{equation}
\mathcal{L}_{bd}(y, \hat{y})=1-\frac{2 P_b^{c} R_b^{c}}{P_b^{c}+R_b^{c}},
\end{equation}
where $P_b^c$ and $R_b^c$ define the precision and recall of predicted boundary image $\hat{y}^{b}$ to real one $y^{b}$ for class $c$. The boundary image is computed as follows:
\begin{equation}
\begin{aligned}
&y^{b}=pool\left(1-y, \theta_{0}\right)-\left(1-y\right) \\
&\hat{y}^{b}=pool\left(1-\hat{y}, \theta_{0}\right)-\left(1-\hat{y}\right)
\end{aligned}
\end{equation}
where $pool(\cdot, \cdot)$ employs a pixel-wise max-pooling. It operates on the inverted ground truth binary map or predictions with a sliding window of size $\theta_{0}=3$.

\section{Experiment}
\label{sec:exp}
In this section, we give the details of our experiments and show the results on LiDAR semantic segmentation. Moreover, we compare our proposed approach against the state-of-the-art methods and discuss the results on different settings. 
\subsection{Datasets}
The SemanticKITTI dataset~\cite{behley2019semantickitti} is a large-scale outdoor scene LiDAR dataset, which provides the complete point-wise labels for all 22 sequences in KITTI Odometry Benchmark~\cite{geiger2013vision}. Sequence 00 to 10 are treated as the training sets, and Sequence 11 to 21 are used as test sets. There are 23,201 and 20,351 complete 3D scans for training and testing, respectively. We follow the setting in~\cite{behley2019semantickitti}, and keep Sequence 08 as the validation set. To evaluate the effectiveness of our proposed approach, we submit the output to the online evaluation website to obtain the results on the testing set without the extra tricks like test time augmentation, fine-tuning on the validation set, or the pre-trained models.

The SemanticPOSS dataset~\cite{pan2020semanticposs} is also collected by the LiDAR scanner in outdoor scene, which is more sparse comparing to SemanticKITTI. It contains 2,988 LiDAR scans captured at the campus with large quantity of dynamic instances. SemanticPOSS is divided into six parts with the same size. As in~\cite{pan2020semanticposs}, we employ the part 3 for evaluation.

\subsection{Evaluation Metric}
\label{sec:eva}
To facilitate the fair comparison, we evaluate the performance of different methods with respect to the mean intersection over union metric (mIoU)~\cite{behley2019semantickitti}, which is defined as below: 
\begin{equation}
            \label{slice}
         	mIoU = \frac{1}{n}\sum_{c=1}^{n}{\frac{TP_c}{TP_c+FP_c+FN_c}}.
\end{equation}
For class $c$, $TP_c$ represents the true positives, and $FP_c$ denotes false positives. $FN_c$ is false negative predictions. 

In SemanticKITTI benchmark, the single scan task evaluates 19 different classes ($n = 19$). On the other hand, the multiple scans task evaluates 25 different classes ($n = 25$), which needs to distinguish more than 6 moving classes comparing to the single-scan challenge. 
In SemanticPOSS benchmark, 11 different classes are evaluated in the single-scan task.

\subsection{Training Settings}
We trained the proposed neural network for $180$ epochs using the stochastic gradient descent (SGD) on a PC with two RTX 2080Ti GPUs. The total batch size is $4$. Moreover, the initial learning rate is set to $0.01$ with a decay of $0.01$ at every epoch. We conducted the inference on a single RTX 2080Ti GPU. The height and width of the range residual image are set to $H = 64$, and $W = 2048$, respectively. During the training process, we perform data augmentation by randomly rotating, transforming, and flipping the 3D point cloud. In addition, we randomly drop the points at a percentage with a uniform distribution between $0$ and $10$ before creating the range residual image.

\subsection{Performance Evaluation}
\begin{table*}%
\begin{center}
 \setlength{\abovecaptionskip}{-0.2cm}
\caption{Comparisons on the SemanticKITTI multiple scans benchmark. The item with arrow indicates the moving class. Values are given as IoU ($\%$). $*$ denotes the FPS measured on a Tesla V100 GPU, while ours are taken on a single RTX 2080Ti GPU.} %
\label{tab:table1}%
\setlength{\tabcolsep}{1.2pt}
\renewcommand{\arraystretch}{1.} 
\begin{tabular}{ c | c | c | c c c c c c c c c c c c c c c c c c c c c c c c c}
  \hline
Methods & \rotatebox{90}{\textbf{mean-IoU}} & \rotatebox{90}{ {\textbf{FPS (Hz)}}} & \rotatebox{90}{car } & \rotatebox{90}{bicycle } & \rotatebox{90}{motorcycle } & \rotatebox{90}{truck } & \rotatebox{90}{other-vehicle } & \rotatebox{90}{person } & \rotatebox{90}{bicyclist } & \rotatebox{90}{motorcyclist } & \rotatebox{90}{road } & \rotatebox{90}{parking } & \rotatebox{90}{sidewalk } & \rotatebox{90}{other-ground } & \rotatebox{90}{building } & \rotatebox{90}{fence } & \rotatebox{90}{vegetation } & \rotatebox{90}{trunk } & \rotatebox{90}{terrain } & \rotatebox{90}{pole } & \rotatebox{90}{traffic sign } & \rotatebox{90}{$\underrightarrow{\text{car}}$} & \rotatebox{90}{$\underrightarrow{\text{bicyclist}}$} & \rotatebox{90}{$\underrightarrow{\text{person}}$} & \rotatebox{90}{$\underrightarrow{\text{motorcyclist}}$} & \rotatebox{90}{$\underrightarrow{\text{other-vehicle}}$ } & \rotatebox{90}{$\underrightarrow{\text{truck}}$} \\
  \hline \hline
  TangentConv~\cite{tatarchenko2018tangent} & 34.1 &  {-} & 84.9 & 2.0 & 18.2 & 21.1 & 18.5 & 1.6 & 0.0 & 0.0 & 83.9 & 38.3 & 64.0 & 15.3 & 85.8 & 49.1 & 79.5 & 43.2 & 56.7 & 36.4 & 31.2 & 40.3 & 1.1 & 6.4 & 1.9 & \textbf{30.1} & \textbf{42.2}\\
  DarkNet53Seg~\cite{behley2019semantickitti}  & 41.6 &  {-} & 84.1 & 30.4 & 32.9 & 20.2 & 20.7 & 7.5 & 0.0 & 0.0 & 91.6 & \textbf{64.9} & 75.3 &  \textbf{27.5}  & 85.2 & 56.5 & 78.4 & 50.7 & 64.8 & 38.1 & 53.3 & 61.5 & 14.1 & 15.2 & 0.2 & 28.9 & 37.8 \\
  SpSequenceNet~\cite{shi2020spsequencenet}  & 43.1 &  {3} & 88.5 & 24.0 & 26.2 & 29.2 & 22.7 & 6.3 & 0.0 & 0.0 & 90.1 & 57.6 & 73.9 & 27.1 & \textbf{91.2} & \textbf{66.8} & 84.0 & 66.0 & 65.7 & 50.8 & 48.7 & 53.2 & 41.2 & 26.2 & 36.2 & 2.3 & 0.1 \\
  TemporalLidarSeg~\cite{duerr2020lidar} & 47.0 &  {30*} & \textbf{92.1} & 47.7 & 40.9 & \textbf{39.2} & \textbf{35.0} & 14.4 & 0.0 & 0.0 & \textbf{91.8} & 59.6 & \textbf{75.8} & 23.2 & 89.8 & 63.8 & 82.3 & 62.5 & 64.7 & 52.6 & 60.4 & 68.2 & 42.8 & 40.4 & 12.9 & 12.4 & 2.1 \\
  
   {TemporalLatticeNet~\cite{schutt2022abstract}}  &  {47.1} &  {6.5} &  {91.6} &  {35.4} &  {36.1} &  {26.9} &  {23.0} &  {9.4} &  {0.0} &  {0.0} &  {91.5} &  {59.3} &  {75.3} &  {\textbf{27.5}} &  {89.6} &  {65.3} &  {\textbf{84.6}}  &  {\textbf{66.7}} &  {\textbf{70.4}} &  {\textbf{57.2}} & {60.4} &  {59.7} &  {41.7} &  {9.4} &  {\textbf{48.8}} &  {5.9} &   {0.0} \\
  \hline 
  Meta-RangeSeg(Ours) &  {\textbf{49.7}} &  {22} &  {90.8} &  {\textbf{50.0}} &  {\textbf{49.5}} &  {29.5} &  {34.8} &  {\textbf{16.6}} &  {0.0} &  {0.0} &  {90.8} &  {62.9}&  {74.8} &  {26.5} &  {89.8} &  {62.1} &  {82.8} &  {65.7} &  {66.5} &  {56.2} &  {\textbf{64.5}} &  {\textbf{69.0}} &  {\textbf{60.4}} &  {\textbf{57.9}} &  {22.0} &  {16.6} &  {2.6} \\

  \hline  
\end{tabular}%
\end{center}
\vspace{-6mm}
\end{table*}%

\begin{table*}
\begin{center}
 \setlength{\abovecaptionskip}{-0.1cm}
    \caption{Comparisons on the SemanticKITTI single scan benchmark. $\ddagger$ denotes the second best results. The top-half shows the point-wise methods, and the bottom-half is projection-based methods. $*$ denotes the results reproduced from the original implementation.}
        \label{tab:table2}
    \setlength{\tabcolsep}{2.pt}
      \renewcommand{\arraystretch}{1.} 
 \begin{tabular}{c| c| c| c | c c c c c c c c c c c c c c c c c c c} 
 \hline
 Methods & Size & \rotatebox{90}{\textbf{mean-IoU}} & \rotatebox{90}{ {\textbf{FPS (Hz)}}} & \rotatebox{90}{car}& \rotatebox{90}{bicycle}& \rotatebox{90}{motorcycle}& \rotatebox{90}{truck}& \rotatebox{90}{other-vehicle}& \rotatebox{90}{person}& \rotatebox{90}{bicyclist}& \rotatebox{90}{motorcyclist}& \rotatebox{90}{road}& \rotatebox{90}{parking}& \rotatebox{90}{sidewalk}& \rotatebox{90}{other-ground}& \rotatebox{90}{building}& \rotatebox{90}{fence}& \rotatebox{90}{vegetation}& \rotatebox{90}{trunk}& \rotatebox{90}{terrain}& \rotatebox{90}{pole}& \rotatebox{90}{traffic-sign} \\ 
 \hline\hline
 PointNet~\cite{qi2017pointnet} & 50K pts &14.6 &  {2} &  46.3 &1.3 &0.3 &0.1& 0.8& 0.2& 0.2& 0.0& 61.6& 15.8& 35.7& 1.4& 41.4& 12.9& 31.0& 4.6& 17.6& 2.4& 3.7\\ 
 PointNet++~\cite{qi2017pointnet++} & 50K pts&20.1&  {0.1} &53.7& 1.9& 0.2& 0.9& 0.2& 0.9& 1.0& 0.0& 72.0& 18.7& 41.8& 5.6& 62.3& 16.9& 46.5& 13.8& 30.0& 6.0 &8.9 \\
 SPLATNet~\cite{su2018splatnet} & 50K pts &22.8&  {1}&66.6 &0.0& 0.0& 0.0& 0.0& 0.0& 0.0& 0.0& 70.4& 0.8& 41.5& 0.0& 68.7& 27.8& 72.3& 35.9 &35.8 &13.8 &0.0\\
 TangentConv~\cite{tatarchenko2018tangent} & 50K pts & 35.9 &  {0.3} &86.8& 1.3& 12.7& 11.6& 10.2& 17.1& 20.2& 0.5 &82.9& 15.2& 61.7& 9.0& 82.8& 44.2& 75.5& 42.5& 55.5& 30.2& 22.2\\
 LatticeNet~\cite{rosu2019latticenet}& 50K pts &52.9&  {7}&92.9 &16.6& 22.2& 26.6& 21.4& 35.6& 43.0& 46.0& 90.0& 59.4& 74.1& 22.0& 88.2& 58.8& 81.7& 63.6& 63.1& 51.9& 48.4\\
  RandLA-Net~\cite{hu2020randla} & 50K pts& 53.9&  {1.3} & 94.2& 26.0& 25.8& 40.1 & 38.9& 49.2 &48.2& 7.2& 90.7& 60.3& 73.7& 20.4& 86.9& 56.3& 81.4& 61.3& 66.8& 49.2& 47.7\\
  KPConv~\cite{thomas2019kpconv} & 50K pts &58.8&  {3.8} &\textbf{96.0}& 30.2& 42.5& 33.4& 44.3 &61.5& 61.6& 11.8& 88.8& 61.3& 72.7& \textbf{31.6}& 90.5 & 64.2& 84.8 & \textbf{69.2}& 69.1& 56.4& 47.4\\
  BAAF-Net~\cite{qiu2021semantic} & 50K pts & 59.9 &  {4.8} & 95.4 & 31.8 & 35.5 & \textbf{48.7} & \textbf{46.7} & 49.5 & 55.7 & 33.0 & 90.9 & 62.2 & 74.4 & 23.6 & 89.8 & 60.8 & 82.7 & 63.4 & 67.9 & 53.7 &52.0 \\ 
 \hline

 RangeNet53++~\cite{milioto2019rangenet++}& $64\times 2048$&52.2&  {12} & 91.4 &25.7& 34.4& 25.7& 23.0& 38.3& 38.8& 4.8& 91.8& 65.0& 75.2& 27.8& 87.4& 58.6& 80.5& 55.1& 64.6& 47.9& 55.9\\
 
 PolarNet~\cite{zhang2020polarnet} &[480, 360, 32] &54.3&  {16}& 93.8 &40.3& 30.1& 22.9& 28.5& 43.2& 40.2& 5.6& 90.8& 61.7& 74.4& 21.7& 90.0& 61.3& 84.0& 65.5& 67.8& 51.8& 57.5\\

 MINet~\cite{li2021multi} & $64\times 2048$& 55.2 &  {24}& 90.1 & 41.8 & 34.0 & 29.9 & 23.6 & 51.4 & 52.4 & 25.0 & 90.5 & 59.0 & 72.6 & 25.8 & 85.6 & 52.3 & 81.1 & 58.1 & 66.1 & 49.0 & 59.9  \\

  3D-MiniNet~\cite{alonso20203d} & $64\times 2048$&55.8 &  {28}&90.5& 42.3& 42.1& 28.5& 29.4& 47.8& 44.1& 14.5& 91.6& 64.2& 74.5& 25.4& 89.4& 60.8& 82.8& 60.8& 66.7& 48.0& 56.6\\
 
  SqueezeSegV3~\cite{xu2020squeezesegv3}& $64\times 2048$&55.9& {6}&92.5& 38.7& 36.5& 29.6& 33.0& 45.6& 46.2& 20.1& 91.7& 63.4& 74.8& 26.4& 89.0& 59.4 &82.0& 58.7& 65.4& 49.6& 58.9\\

  {CNN-LSTM~\cite{wen2022hybrid}} &  {[512, 512, 32]} &   {56.9} &  {11} &  {92.6} &  {45.7}&  {49.6}&  {48.6}&  {30.2}&  {53.8}&  {\textbf{74.6}}&  {9.2}&  {90.7}&  {23.3}&  {75.7}&  {17.6}&  {90.0}&  {51.3}&  {\textbf{87.1}}&  {60.8}&  {\textbf{75.4}}&  {\textbf{63.9}}&  {41.5}\\

  SalsaNext~\cite{cortinhal2020salsanext}& $64\times 2048$&59.5&  {24}& 91.9& 48.3& 38.6& 38.9& 31.9& 60.2& 59.0& 19.4& 91.7& 63.7& 75.8& 29.1& 90.2& 64.2 & 81.8& 63.6& 66.5 &54.3& 62.1 \\
  
   FIDNet~\cite{zhao2021fidnet} & $64\times 2048$&59.5&  {29*}&93.9&\textbf{54.7}&48.9&27.6&23.9&62.3&59.8&23.7&90.6&59.1&75.8&26.7&88.9&60.5&84.5&64.4&69.0&53.3&62.8\\
   Lite-HDSeg~\cite{razani2021lite} & $64\times 2048$&\textbf{63.8}&  {20}&92.3&40.0&\textbf{55.4}&37.7&39.6&59.2&71.6&\textbf{54.1}&\textbf{93.0}&\textbf{68.2}&\textbf{78.3}&29.3&\textbf{91.5}&\textbf{65.0}&78.2&65.8&65.1&59.5&\textbf{67.7}\\

 \hline
 
 Meta-RangeSeg(Ours) & $64\times 2048$&61.0$\ddagger$&  {26} &93.9& 50.1 &43.8&43.9 &43.2&\textbf{63.7}&53.1&18.7&90.6&64.3&74.6&29.2&91.1&64.7&82.6&65.5&65.5&56.3&64.2\\
 \hline
\end{tabular}
\end{center}
    \vspace{-6mm}
\end{table*}

For the quantitative evaluation, we compare our proposed approach against the previous methods on the multiple scans semantic segmentation benchmark, which is our main focus. As shown in Table~\ref{tab:table1}, our presented Meta-RangeSeg method outperforms the state-of-the-art approach~\cite{schutt2022abstract} by 2.6\% on SemanticKITTI testing set without extra tricks.
It is worthy of mentioning that our approach can process the multiple scans point clouds at a rate of 22Hz while maintaining high accuracy. It is faster than the frame rate (10Hz) of the sensor used in SemanticKITTI dataset.  

Considering that there are few methods in the multiple scans evaluation, we conduct experiments on the single scan task of SemanticKITTI and SemanticPOSS to illustrate the generalization capability of our proposed approach. For a fair comparison, we only use the range image by excluding the residual image when extracting features under range view for the single scan evaluation. As shown in Table~\ref{tab:table2} and Table~\ref{tab:table3}, Meta-RangeSeg outperforms most of previous methods under the same settings, including point-wise and projection-based methods, which demonstrates the effectiveness of our proposed approach on LiDAR semantic segmentation. 

For the qualitative evaluation\footnote{Supplementary video: \url{https://youtu.be/xUFsmmjZYuA}}, Fig. \ref{fig:sixth} shows a sample of semantic segmentation results on the SemanticKITTI validation set. The results under range view are generated by our method with the complete semantic labels. To investigate the performance of our approach on multiple scans semantic segmentation, we take the car class including static and moving as example to compare the predicted visual results with the ground truth and predictions generated by SpsequenceNet \cite{shi2020spsequencenet}, as illustrated in Fig.~\ref{fig:seventh}. It can be observed that our approach can effectively distinguish both static and moving objects with their semantic information. 

\begin{table}[h]
\vspace{-2mm}
\begin{center}
\setlength{\abovecaptionskip}{-0.1cm}
\caption{Comparisons on the SemanticPOSS benchmark. Values are given as IoU ($\%$). $*$ denotes the results reproduced from the original implementation.}
\label{tab:table3}%
\setlength{\tabcolsep}{4pt}
\renewcommand{\arraystretch}{1.} 
\scalebox{0.7}{
\begin{tabular}{ c | c | c c c c c c c c c c c c c c c c c c c c c c c c c}
  \hline
{Methods} & \rotatebox{90}{ {\textbf{mean-IoU}}}  & \rotatebox{90}{ {people}} & \rotatebox{90}{ {rider}} & \rotatebox{90}{ {car}} & \rotatebox{90}{ {traffic sign}} & \rotatebox{90}{ {trunk}} & \rotatebox{90}{ {plants}} & \rotatebox{90}{ {pole}} & \rotatebox{90}{ {fence}} & \rotatebox{90}{ {building}} & \rotatebox{90}{ {bike}} & \rotatebox{90}{ {road}}  \\
  \hline \hline
   {SqueezeSegV2~\cite{wu2019squeezesegv2}}  &  {29.8} &  {18.4} &  {11.2}  &  {34.9} &  {11.0} &  {15.8} &  {56.3} &  {4.5} &  {25.5} &  {47.0}  &  {32.4}  &  {71.3} \\
   {RangeNet53++~\cite{milioto2019rangenet++}}  &  {28.9} &  {14.2} &  {8.2} &  {35.4} &  {6.8} &  {9.2} &  {58.1} &  {2.8} &  {28.8} &  {55.5} &  {32.2} &  {66.3}  \\
   {UnpNet~\cite{li2021rethinking}}  &  {34.3} &  {17.7} &  {17.2} &  {39.2} &  {9.5} &  {13.8} &  {67.0} &  {5.8} &  {31.1} &  {66.9} &  {40.5} &  {68.4}  \\
   {MINet~\cite{li2021multi}} & {35.1}  &  {20.1} &  {15.1} &  {36.0} &  {15.5} &  {23.4} &  {67.4} &  {5.1} &  {28.2} &  {61.6} &  {40.2} &  {72.9} \\
   {SalsaNext~\cite{cortinhal2020salsanext}}  &  {49.4*} &  {73.4} &  {17.1} &  {73.3} &  {7.0} &  {25.5} &  {69.1} &  {26.9} &  {45.0} &  {77.1} &  {50.1} &  {\textbf{79.3}}  \\
  \hline 
   {Meta-RangeSeg(Ours)} &  {\textbf{53.7}} &  {\textbf{75.5}} &  {\textbf{19.8}} &  {\textbf{78.7}} &  {\textbf{25.6}} &  {\textbf{27.5}} &  {\textbf{72.3}} &  {\textbf{32.3}} &  {\textbf{49.0}} &  {\textbf{78.0}} &  {\textbf{52.9}} &  {\textbf{79.3}} \\
  \hline  
\end{tabular}}%
\end{center}
\vspace{-6mm}
\end{table}%

\begin{figure*}
\centering
\setlength{\abovecaptionskip}{-0.1cm}
\includegraphics[width=0.9\linewidth,height=45mm]{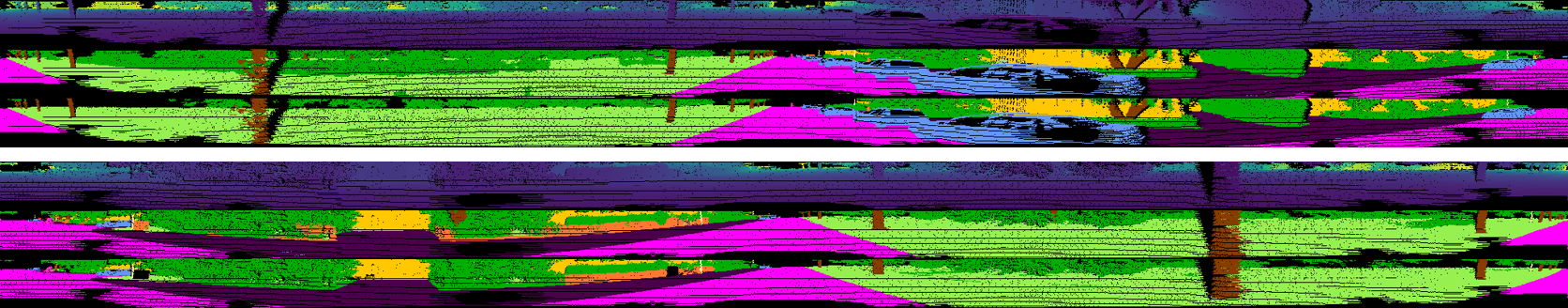}
    \caption{ 
    Semantic segmentation results from the range view on the SemanticKITTI validation set (sequence 08) [best view in color].  In each scene, we show the range channel image, predictions by Meta-RangeSeg, and ground truth in turn.}
    \label{fig:sixth}
    \vspace{-4mm}
\end{figure*}

\begin{figure*}
\setlength{\abovecaptionskip}{-0.3cm}
 \centering
\includegraphics[width=0.9 \linewidth,height=69mm]{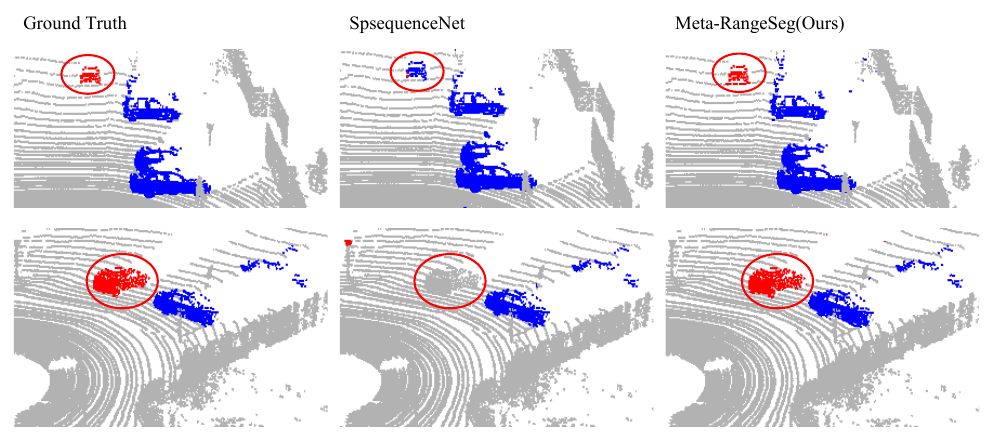}
    \caption{Segmentation results of static and moving cars on the SemanticKITTI validation set (sequence 08). The static cars are labeled in blue, and moving cars are in red [best view in color]. At each row, the left is the ground truth, the middle is the predictions by SpsquencesNet~\cite{shi2020spsequencenet}, and the right is the results of our proposed Meta-RangeSeg approach.}
    \label{fig:seventh}
    \vspace{-4mm}
\end{figure*}

\subsection{Ablation Study}
To examine the improvements of each individual module in our proposed network, we conduct the ablation studies on Sub-SemanticKITTI dataset, which is a subset of the original SemanticKITTI. The training set of Sub-SemanticKITTI dataset consists of every 8th frame in sequence 00-10~(except 08). The validation set is formed by every 4th frame in sequence 08. As only 1/8 training data and 1/4 validation data are used, we can quickly perform the evaluation. 

\vspace{-4mm}
\begin{table}[h]
\begin{center}
\setlength{\abovecaptionskip}{-0.1cm}
\caption{Ablative analysis evaluated on validation set (seq 08) in Sub-SemanticKITTI dataset. }
\label{tab:table4}
\scalebox{0.7}{
\begin{tabular}{|c|ccccc|c | c| c| }
\hline 
\multicolumn{1}{|c|}{\textbf{Architecture}} &
\multicolumn{1}{c}{\textbf{R$^2$ Image}} & 
\multicolumn{1}{c}{\textbf{Meta-Kernel}} &
\multicolumn{1}{c}{\textbf{FAM}} &
\multicolumn{1}{c|}{\textbf{\begin{tabular}[c]{@{}c@{}}Boundary\\ Loss\end{tabular}}} &
\multicolumn{1}{c|}{\textbf{mIoU}} &
\multicolumn{1}{c|}{ {\textbf{Params}} } &
\multicolumn{1}{c|}{ {\textbf{FLOPs}}}\\
 \hline 
\multirow{1}{*}{Baseline}
 & \xmark & \xmark & \xmark & \multicolumn{1}{c|}{\xmark}  & 42.8 &  {6.71M} &  {125.74G}  \rule{0pt}{3ex}\\ \hline
 
\multirow{9.5}{*}{Meta-RangeSeg}

 & \checkmark & \xmark & \xmark & \multicolumn{1}{c|}{\xmark}  & 43.3 &  {6.71M} &  {125.76G}  \rule{0pt}{3ex}\\ \cline{2-8}  

 & \checkmark &\checkmark   & \xmark & \multicolumn{1}{c|}{\xmark}  & 44.2  &  {6.66M} &  {115.70G} \rule{0pt}{3ex}\\ \cline{2-8}  

 & \checkmark &\checkmark  & \checkmark   & \multicolumn{1}{c|}{\xmark}  & 45.1 &  {6.79M} &  {148.64G}  \rule{0pt}{3ex}\\ \cline{2-8}  

 &  {\xmark} &  {\checkmark}  &  {\checkmark}   & \multicolumn{1}{c|}{ {\checkmark}}  &  {45.3} &  {6.78M}&  {147.30G}  \rule{0pt}{3ex}\\ \cline{2-8}  
 
 &  {\checkmark} & {\xmark}  &  {\checkmark}   & \multicolumn{1}{c|}{ {\checkmark}}  &  {44.2}  &  {6.84M}&  {158.70G} \rule{0pt}{3ex}\\ \cline{2-8}  
 
 &  {\checkmark} &  {\checkmark}  &  {\xmark}   & \multicolumn{1}{c|}{ {\checkmark}}  &  {44.5} &  {6.66M} &  {115.70G}  \rule{0pt}{3ex}\\ \cline{2-8}  

 & \checkmark &\checkmark  & \checkmark   & \multicolumn{1}{c|}{\checkmark}  & 46.9  & {6.79M} &  {148.64G} \rule{0pt}{3ex}\\ \hline

\end{tabular}}
\end{center}
\vspace{-10mm}
\end{table}

For fair comparison, we treat SalsaNext~\cite{cortinhal2020salsanext} with a 5-channel range image as the baseline method, which uses the similar backbone network as ours. In our experiments, we report the results on multiple scans semantic segmentation task. As shown in Table \ref{tab:table4}, our proposed range residual image (R$^2$ Image) with 9 channels outperforms the baseline around 0.5\% , which demonstrates that the range residual is effective. Moreover, the Meta-kernel block obtains over 0.9\% improvement comparing the method without it. Furthermore, FAM performs better than the network without it around 0.9\%. Additionally, the boundary loss achieves over 1.8\% performance gain, which indicates that the boundary regions are essential to the LiDAR semantic segmentation. 
Then, we remove each module individually from the complete framework while the results in accuracy drop by 1.6\%-2.7\%, separately.
Finally, it can be observed that our proposed Meta-RangeSeg approach outperforms the baseline over 4.1\% with only 1.2\% (i.e., 0.08M) extra parameters, which demonstrates the efficacy of each module.  
Although having added the extra blocks into the backbone, the inference speed of the final model is not greatly affected. This is mainly due to the concise input and our proposed lightweight modules.

\section{Conclusion}
\label{sec:conclusion}

This paper proposed a novel approach to LiDAR semantic segmentation, which introduced a range residual image representation to capture the spatial-temporal information. Moreover, we employed Meta-Kernel to extract the meta features from the residual image by dynamically learning the weights from the relative Cartesian coordinates and range values. Furthermore, we designed a Feature Aggregation Module to aggregate the features at various scales and levels while emphasizing the range channel. We have conducted extensive evaluations on SemanticKITTI and SemanticPOSS benchmark, whose promising results demonstrated that our proposed Meta-RangeSeg approach not only outperforms the state-of-the-art semantic segmentation methods on multiple scans benchmark but also runs at 22 FPS. 

Since some modules are not fully optimized, our model consumes more memory. In future, we will try to design an effective model that can make use of the connections between each channel in range residual image.


\bibliographystyle{IEEEtran}
\bibliography{IEEEabrv,IEEEexample}


 





\end{document}